\documentclass{article}

\usepackage{arxiv}

\usepackage[utf8]{inputenc} % allow utf-8 input
\usepackage[T1]{fontenc}    % use 8-bit T1 fonts
\usepackage{hyperref}       % hyperlinks
\usepackage{url}            % simple URL typesetting
\usepackage{booktabs}       % professional-quality tables
\usepackage{amsfonts}       % blackboard math symbols
\usepackage{nicefrac}       % compact symbols for 1/2, etc.
\usepackage{microtype}      % microtypography
\usepackage{lipsum}
\usepackage{graphicx}
\usepackage{subfigure}
\usepackage{amsmath}
\usepackage{times}
\usepackage{amssymb}
\usepackage{mathptmx}
\usepackage{float}
\usepackage{comment}
\usepackage[usenames,dvipsnames]{xcolor}
\include{localdefs}
\usepackage{color}
\usepackage{etoolbox}

\title{Object Detection based on LIDAR Temporal Pulses using Spiking Neural Networks}

\author{
  Shibo Zhou \\
  Department of Electrical and Computer Engineering\\
  Binghamton University\\
  The State University of New York\\
  Binghamton, NY 13902 \\
  \texttt{szhou19@binghamton.edu} \\
   \And
 Wei Wang \\
  Department of Computer Science and Engineering\\
  University at Buffalo\\
  The State University of New York\\
  Buffalo, NY 14260 \\
  \texttt{wwang49@buffalo.edu} \\
}

\begin{document}
\maketitle

\begin{abstract}
Neural networks has been successfully used in the processing of Lidar data, especially in the scenario of autonomous driving. However, existing methods heavily rely on pre-processing of the pulse signals derived from Lidar sensors and therefore result in high computational overhead and considerable latency. In this paper, we proposed an approach utilizing Spiking Neural Network (SNN) to address the object recognition problem directly with raw temporal pulses. To help with the evaluation and benchmarking, a comprehensive temporal pulses data-set was created to simulate Lidar reflection in different road scenarios. Being tested with regard to recognition accuracy and time efficiency under different noise conditions, our proposed method shows remarkable performance with the inference accuracy up to 99.83\%\ (with 10\%\ noise) and the average recognition delay as low as 265 ns. It highlights the potential of SNN in autonomous driving and some related applications. In particular, to our best knowledge, this is the first attempt to use SNN to directly perform object recognition on raw Lidar temporal pulses.
\end{abstract}

% keywords can be removed

\section{Introduction}
%As is known to all, a difficult problem in machine vision is to judge the distance of object. 
In computer vision for autonomous driving, one of the most difficult problems is to detect objects in three-dimensional space with low delay at long distance \cite{hwu2018adaptive}.
Traditional method for 2D image processing usually perform poorly in this task as a single camera could not obtain accurate depth information \cite{lu2006single}. 
Generating depth map based on multi-camera can improve depth accuracy, however, its high computing complexity prevents it to meet the real-time requirement \cite{park1998acquisition}. 
%Another awkward problem is that the optics camera is subject to great lighting conditions and the object’s accuracy is uncertain. 
%Another awkward problem for optics camera is its dependency on lighting conditions, which results in uncertain and unstable performance in extreme weather conditions.
Radio detection and ranging (Radar) is immune to lighting variations as it uses radio waves to determine the range, angle, or velocity of objects. 
However, its short wavelength properties do not allow the detection of small objects nor provide users precise object images \cite{difranco2004radar}. 
%LIDAR with short-wavelength and full-angle field characteristics can solve the above problems to a large extent. By means of the nature of LIDAR, LIDAR are becoming a reliable solution for high speed and precision object detection.
By using laser signal with relatively shorter wavelength than radio waves, Light detection and ranging (Lidar) system provides better range, higher spatial resolution and a larger field of view than Radar to help detect obstacles on the curves \cite{weitkamp2005lidar}.

Comparing to other sensors, Lidar has significant advantage in detecting and recognizing objects at long distance in a wide view, so that self-driving cars at high speed can take evasive actions in time.
      
Although significant amount of research has been done for object classification from 3-D point clouds, It remains an open question on how to do object detection and recognition with raw Lidar temporal pulses.
Lidar uses active sensors which emit their own energy source for illumination, and detects/measures the reflected energy from the objects. 
To analyze sensor output, traditional Lidar system needs multi-stages of signal processing, including analog to digital conversion, averaging and filtering, in order to produce high quality digital 3-D point clouds for later object detection and recognition by different learning algorithms, such as traditional feature extraction method \cite{behley2013laser, wang2015voting, gonzalez2017board, tatoglu2012point, hernandez2014lane} and novel neural network method \cite{li2016vehicle, li20173d, chen2017multi, oh2017object, kim2016robust, asvadi2017depthcn, asvadideep}. 
The signal processing flow between Lidar sensor and object detection consumes high computing power and causes unignorable delays, it is important to find a solution which can realize object detection and recognition based on analog temporal pulses from Lidar sensors.   
 
Spiking Neural Network (SNN), catches our attention as it can directly take temporal pulses as input and output classification result. 
Moreover, SNN has a special advantage over regular neural networks on processing temporal information \cite{ponulak2011introduction}:
%in speed, energy efficiency and computational complexity  
%Before the information can spreed to the next layer, every neuron in layer need to be evaluated due to the nature of regular neural networks synchronization processing information. 
For regular neural networks, all neurons are synchronized, so every neuron in the same layer needs to be evaluated before information can be passed to the next layer. 
On the contrary, SNN processes information in an event-based manner, which is asynchronous. 
Here event-based processing has several advantages, 
first of all, when the neuron is addressed by an event, only this neuron is activated, allowing SNN to become much more energy efficient \cite{susi2018fns}. Second, the event can be responded directly by a neuron and does not have to wait until all the neurons in layer are evaluated, nor to the next discrete time step to fire its response. It shows that the ability of SNN to process information without delay \cite{martin2015spiking}. Third, the event can be processed using a relatively small number of spikes, reducing the computational complexity \cite{rasshofer2005automotive}. In a word, SNN can implement real-time pulse signal processing. Based on the advantage of SNN and spike nature of itself, SNN could perfectly process the Lidar temporal pulse signal.

%In this paper, we introduce an object detection system based on temporal pulses with different delay and SNN classifiers.
In this paper, we explore the usage of Spiking Neural Network to perform object detection and recognition based on raw temporal pulses from Lidar sensors.  
%The temporal pulses with different delay are produced from Lidar single photon detector array.
In our simulations, Lidar system fires laser signals to the front of vehicle at certain frequency.
Raw temporal pulses are generated by Lidar single photon detector array when it receives reflected photons from the object. 
In this way, depth information is recorded by the different delay of temporal pulses. 
Spiking neural network circuit takes temporal pulses as input, and directly output classification result.  

%Temporal pulses run through the trained spiking-based SNN to achieve object detection.
Our contributions are:
\begin{itemize}
\item We created a complete temporal pulses datasets of Lidar with different road conditions and target objects in different noise environment. Although the Lidar datasets already exists, these datasets are preprocessed through DSP, such as Udacity and KITTI datasets \cite{geiger2013vision}. Our datasets directly simulate the Lidar's temporal pulses signal, so it not only meets the requirements of the simulations but also has practical significance.
\item We firstly applied SNN to directly process raw temporal pulses signal from Lidar sensors. 
%It includes the SNN, a directly implementation of temporal coding where information is encoded in spike times instead of spike rates. \hm{This sentence is meaningless}
We carefully studied existing SNN models and adapt an adequate model for this task. The temporal pulses signal from Lidar do not require any preprocessing and are directly used as input of SNN.
\item The performance of the SNN-based object detection system is carefully evaluated under different noise conditions and show high accuracy as well as ultra-low delay. The simulation results show inference accuracy up to 99.83\%\ with 10\%\ noise, and the average recognition delay is only 265 ns after the first temporal pulse arrives.

\end{itemize}

The paper structure is organized as follows. Section \uppercase\expandafter{\romannumeral2} introduces the background and related works. Section \uppercase\expandafter{\romannumeral3} details Lidar temporal pulses and SNN with temporal coding. The detection performance of the whole system is evaluated and analysis in section \uppercase\expandafter{\romannumeral4}, and in section \uppercase\expandafter{\romannumeral5} we conclude the paper.

\section{Background}

\subsection{Object detection approaches using 3D-Lidar}

\begin{figure}[ht]
\centering
\includegraphics[width=1\linewidth]{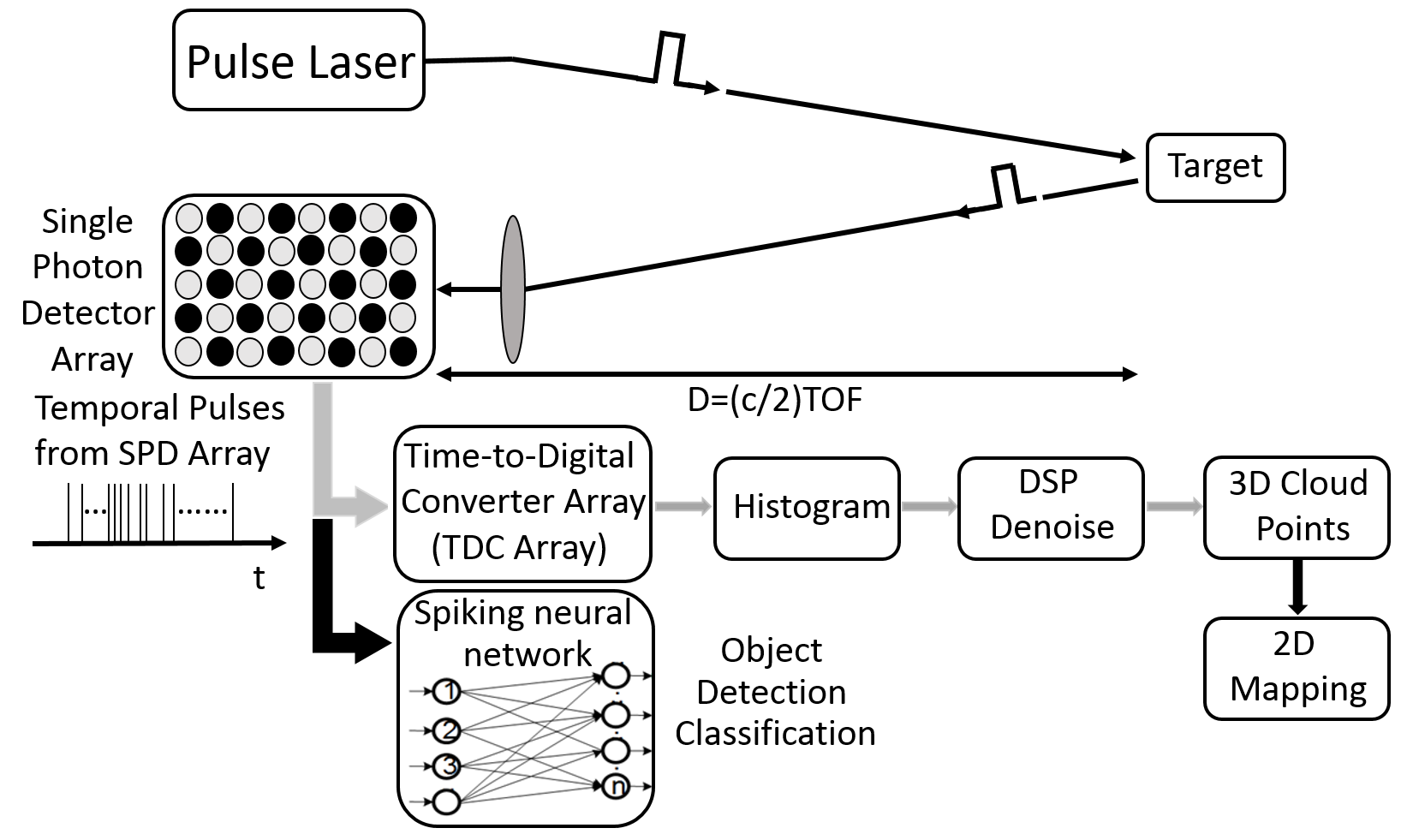}
\caption{Current pattern recognition on Lidar and Our proposed approach.}
\label{fig:lidar}
\end{figure}

In recent years, Lidar is beginning to gain people's attention due to its high resolution and 3D monochromatic image on the object. 
%Firstly, a simple of overview of how the Lidar works.
Generally, it works in the following manner:
The device emits laser pulses which move outwards in various directions until the signals reach an object, and then reflect and return to the receiver. At the same time, an  inner processor saves each reflection points of a laser and generates a 3D cloud points of the environment. Furthermore, the time interval between pulses leaving the device and their returning to Lidar sensors are measured through the same processor. The distance between a detected object and a Lidar receiver can be determined from the running time. However, this whole process requires a series of transformations as illustrated in the Figure \ref{fig:lidar}. When the signal return to the receiver, the detector will generates a single photon detector array (SPDA) and records different temporal pulses information. Next, the time corresponding to the temporal pulses needs to be converted into a digital form followed by histogram, DSP denoise. Finally, the 3D cloud points are generated. In some cases, the 3D cloud points need to be further processed such as the mapping of 3D to 2D. Based on the 3D cloud points from different objects, the poster-processing system can differentiate objects. In addition to using 3D cloud points to detect object, sometimes the intensity of return light can also be used as a method to detect object. The light intensity is directly related to the reflectivity of the object, so the light intensity can also be used as an object detection, but the same transformation is required before the light intensity is entered into the poster-processing system. After transformation, how to detect objects quickly and accurately, the poster-processing system plays an important role in the detection of objects. Now, multiple poster-processing system have been designed to execute object detection. These poster-processing systems have followed one of the two approaches. One way is based on traditional approach such as hierarchical segmentation, sliding windows. the other way is a combination of neural networks. More detail please refer to section III related work.

\subsection{Spiking Neural Network model}
% \begin{figure}
% 		\begin{subfigure}[t]{0.2\textwidth}
% 			\includegraphics[scale=0.5]{fig/SNN2.png}
%         	\caption{SNN}
%         	\label{fig}
% 		\end{subfigure}
%         \hspace{0.4in}
%         ~
%         \begin{subfigure}[t]{0.2\textwidth}
% 			\includegraphics[scale=0.5]{fig/neuron1.png}
%         	\caption{Neuron}
%         	\label{fig}
% 		\end{subfigure}
%         \caption{(a) A 3-layer SNN; (b) A Neuron}
%         \label{fig}
% \end{figure}

\begin{figure}
    \centering
    \subfigure[SNN]{
        \includegraphics[scale=0.5]{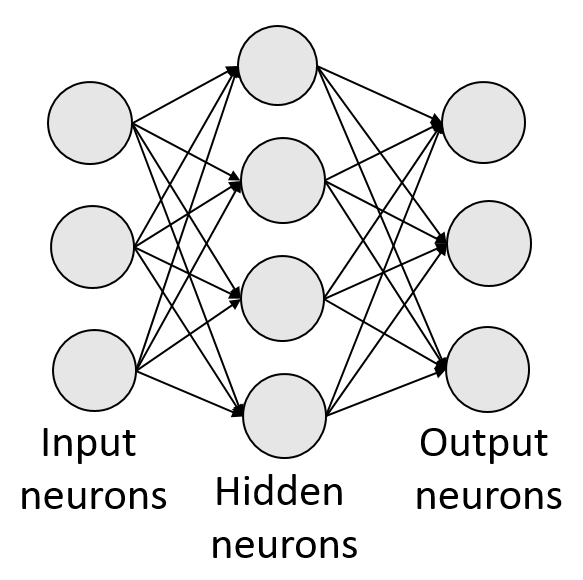}
        \label{fig:snn}}
    \hspace{0.4in}
    \subfigure[Neuron]{
        \includegraphics[scale=0.5]{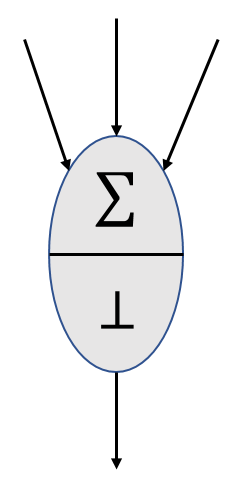}
        \label{fig:neuron}}
    \caption{(a) A 3-layer SNN; (b) A Neuron}
\end{figure}

%The artificial neural network (ANN) become more and more popular with the powerful computational ability in function estimation, complex pattern recognition, and classification. 
Spiking Neural Network (SNN) is regarded as the third generation of Artificial Neural Network (ANN) \cite{maass1997networks} and it has been successfully applied to different domains.
%Throughout their development, the ANN has evolved into the third generation: S (SNN) . 
These applications include pattern generator and controller for different biological model of neuro-prosthetics system \cite{ponulak2006resume}, adaptive robot path planning \cite{hwu2018adaptive}, obstacle recognition and avoidance by modeling and classifying spatio-temporal video data \cite{ge2017spiking}, financial data forecasting based on Polychronous Spiking Network in a unsupervised learning manner \cite{reid2014financial}, composer classification of a musical composition \cite{prasad2015composer}, deal with spatio- and spectro-temporal brain data \cite{kasabov2014neucube},and real time gait-event detection in a supervised learning manner \cite{hwu2018adaptive}. 
At the same time, ROLLS microprocessor along with a Dynamic Vision Sensor \cite{qiao2015reconfigurable} and TrueNorth chip \cite{merolla2014million} has demonstrated superior performance in detection and classification. 

\begin{figure}
\centering
\label{fig:vmem}
\includegraphics[width=1\linewidth]{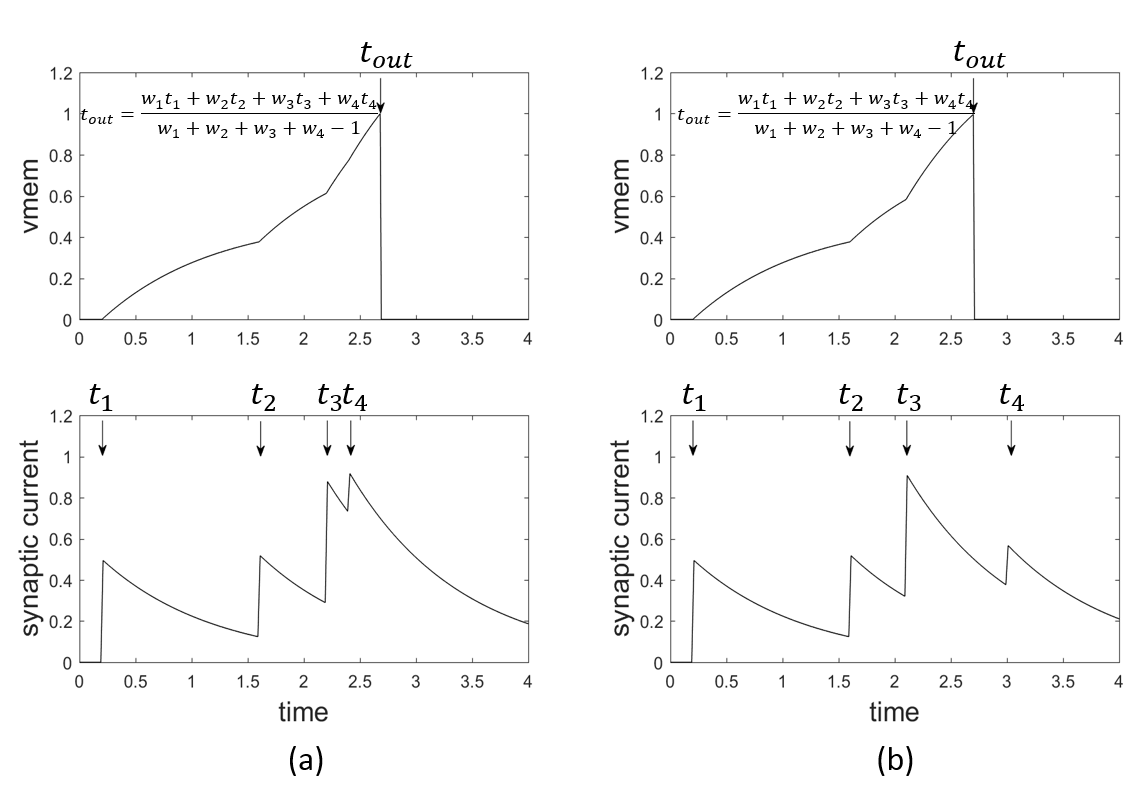}
\caption{The working principle of model in two situations. In (a), the membrane voltage potential of neuron cell reaches the threshold and fire at time $t_{out}$ after receiving 4 spikes with weights $\left \{ w_{1}, w_{2}, w_{3}, w_{4}\right \}$ at the times $\left \{ t_{1}, t_{2}, t_{3}, t_{4}\right \}$. In (b), the membrane voltage potential of neuron cell reaches the threshold and fire before the fourth input spikes arrives, which contrasts sharply with (a). One more thing, a neuron is only allowed to spike once, unless the network is reset or a new input pattern is present.} 
\end{figure}

In SNNs, neurons communicate with spikes or actions potentials through layers of network. When the membrane potential of a neuron reaches to its firing threshold, the neuron will fire a spike to other connected neurons. 
SNN topologies can be classified into three general categories: feedforward networks, recurrent networks and hybrid networks \cite{ponulak2011introduction}. 
The SNN topology used in this work is feedforward networks. Figure \ref{fig:snn} shows an example of fully-connected feedforward SNN, it includes three layers as input layer, hidden layer, and output layer. Number of hidden layers can be more than one for more powerful and complex neural networks. 
Figure \ref{fig:neuron} illustrates a model of spike neuron, which involves accumulation operation and threshold comparative. 
In this work, we use non-leaky integrate and fire (n-LIF) neurons model with exponentially decaying synaptic current kernels \cite{mostafa2018supervised}.
The neuron's membrane dynamic is described by:
\begin{equation}
\centering
\frac{dV^{j}_{mem}(t)}{dt} = \sum\limits_{i}\omega _{ji}\sum\limits_{r}\kappa(t-t^{r}_{i})
\end{equation}
Where $V_{mem}^{j}$ is the membrane potential of neuron j. $w_{ji}$ is the weight of the synaptic connection from neuron i to neuron j and $t_i^r$ is the time of the $r^{th}$ spike from neuron i. 
$\kappa$ is the synaptic current kernel function as given below:
\begin{equation}
\centering
\kappa(x)=\Theta(x)\exp(-\frac{x}{\tau_{syn}}),\  \text{where} \ \Theta(x) = 
\begin{cases}
1 & \text{if $x\geq0$}\\
0 & \text{otherwise}
\end{cases}
\end{equation}

% \begin{equation}
%     \kappa (x)=\Theta (x)\exp(-\frac{x}{\tau _{syn}}), \ \mbox{where}\  \Theta (x)=\left\{\begin{case}
%     1  \mbox{if} \ x \geq 0  \\ 
%     0  \mbox{otherwise} 
%     \end{case}\right.
% \end{equation}

where $\tau_{syn}$ is the only time constant. When the neuron receives a spike, the synaptic current will jump instantaneously, then decays exponentially with time constant $\tau_{syn}$. Figure \ref{fig:vmem} shows how this SNN works. The spike is transformed by the synaptic current kernel and a neuron is only allowed to spike once. The linear relation between input and output spike will be different due to the set of causal input spikes changes \cite{mostafa2018supervised}.

\section{Related work}

This section gives a concise overview of 3D-Lidar-based object detection methods, including traditional methods and neural network methods. The so-called traditional method refers to the realization of object detection using the framework of sliding window or the technique of segmentation, mainly including three steps: the first step is to use the sliding window or segmentation to locate the region of interest as the candidate. Next, extract visual features related to candidate area. Finally, use classifier for identification. As the traditional object detection methods encounter bottlenecks, object detection based on deep leaning is developed. So the method of neural network appears.

For traditional approaches, Behley et al. \cite{behley2013laser} propose a hierarchical segmentation of laser range data approach to realize object detection. Wang and Posner \cite{wang2015voting} applying the voting scheme to process Lidar range data and reflectance values to enable 3D object detection. Gonzales et al. \cite{gonzalez2017board} explore the fusion of RGB and Lidar-based depth maps (front view) by high-definition light detection and ranging. Tatoglu and Pochiraju \cite{tatoglu2012point} presented techniques to model the intensity of the laser reflection return from a point during Lidar scanning to determine diffuse and specular reflection properties of the scanned surface. Hernandez et al. \cite{hernandez2014lane} taking advantage of the reflection of the laser beam identify lane markings on a road surface. 

Recently, people begin to apply neural networks to process Lidar data as they show high accuracy and low latency. Li et al. \cite{li2016vehicle} used a single 2D end-to-end fully-Convolutional Network in a 2D point map from projection of 3D-Lidar range data. Li \cite{li20173d} extended the fully convolutional network based detection techniques to 3D and processed the 3D boxes from Lidar point cloud data. Chen et al. \cite{chen2017multi} provided a top view representation of point cloud from Lidar range data, and combine it with ConvNet-based fusion network for 3D object detection.  Oh et al. \cite{oh2017object} used one of the two independent ConvNet-based classifiers in the depth map from Lidar point cloud. Kim and Ghosh \cite{kim2016robust} proposed a framework using Fast R-CNN to integrate Lidar range data to improve the detection of regions of interest and subsequent identification of road user. Asvadi et al. \cite{asvadi2017depthcn} introduced Deep Concolutional Neural Network to process 3D-Lidar point cloud to predict 2D Bounding Boxes at proposal phase. Asvadi et al. \cite{asvadideep} used deep convolutional neural network to process a dense reflection Map which is generated from 3D-Lidar reflection intensity for object detection.

However, for all aforementioned methods, object detection systems operate on range data or reflection intensity but not raw Lidar signals. 
As showed in Figure \ref{fig:lidar}, all existing methods require heavy pre-processing of Lidar signal before they can be used for object detection or other purposes, which introduces unignorable delay in applications. 
In contrast, our SNN features spike communications, and can directly process temporal pulses from Lidar sensors to implement object detection and recognition. 
%Due to the temporal pulse with different delay contain object information, SNN could directly decide what are the obstacles and where are the obstacles. \hm{redundant} 
%In the future, we could explore to fuse with RGB image-based object detector to improve the overall performance. \hm{makes non sense to put future work here}

\section{Object detection with Spiking Neural Network}

\begin{figure}[ht]
\centering
\label{fig:pipeline}
\includegraphics[width=1\linewidth]{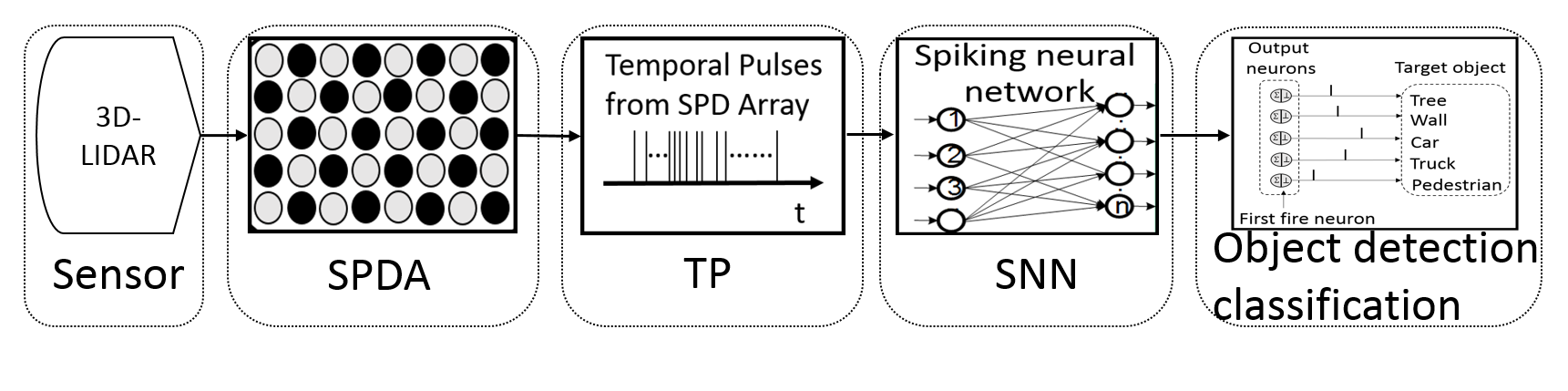}
\caption{The pipeline of the SNN-based object detection system.}
\end{figure}

Figure \ref{fig:pipeline} shows the architecture of our proposed SNN-based object detection system for Lidar signal. As shown in Figure \ref{fig:lidar} the laser emits a pulse which is reflected by the target, and after one photon reaches the detector. The SNN could directly process the temporal pulses from raw SPDA array thus obtain high precision and faster target recognition. This architecture generally consists of three parts: 1.	single photon detector array's (SPDA) temporal pulses contain object information. 2. SNN with temporal coding process directly the temporal pulses signal. 3. The SNN model provides reliable computations and implement object detection and classification.

\subsection{Spiking Neural Network with Temporal Coding}

The different delayed temporal pulses from Lidar’s single photon detector array contain object information. 
For different objects, the temporal pulses have different delay sequences. 
If this sequence can be directly simulated by a coding form, the pulse signal processing can achieve lower latency and higher accuracy performance. 
Therefore, how to implement simulation through coding, plays a crucial role in our research.

We develop a simulation method that SNN with temporal coding.  Although there are multiple coding form based on SNN, our proposed temporal coding has some unique advantages. The following is a detail introduction. Firstly, for computer simulation, the pulse cannot become the input, we need to find a value that corresponding to pulse as input and output of network. In most cases, the value is spike counts or spike rates at a particular time window, but there are some problems. The spike counts are discrete and training phase is a great challenge. To avoid the above problem, we chose the continuous and more precision spike times as the information-carrying quantities. Thus, the spike times is not only relate to temporal pulses but also used as a coding form of SNN. Secondly, For the temporal coding, even though some algorithms have been proposed, the algorithms still has some drawbacks. Such as the SpikeProp algorithms \cite{bohte2002error}, which described the cost function in term of the difference between desired and actural spike times, was limited to learning a single spike. Supervised Hebbian learning \cite{legenstein2005can}, and ReSuMe \cite{ponulak2010supervised}, they primarily suitable for training in single-layer networks, cannot perform more complex computation. To address this problem, we develop a conventional network model that relies only on simple neural and synaptic dynamics instead of complex and discontinuous dynamics of spiking neurons. If so, we not only avoid the discreteness of spike, but could extend naturally to multi-layer networks. Therefore, the temporal coding we proposed can realize SNN directly processing temporal pulses signal from Lidar and active function derived from the non-leaky integrate and fire neurons \cite{mostafa2018supervised}.

\begin{equation}
\centering
exp(t_{out}) = \frac{\substack{\sum_{i\in C}\omega_{i}exp(t_{i})}}{\substack{\sum_{i\in C}\omega_{i}}-threshold}
\end{equation}
Where $t_{out}$ is the neuron’s response time of next layer. 
$t_{i}$ is the firing time of i-th source neuron. $w_{i}$ is the weight corresponding to i-th source neuron. $C=\left \{ i: t_{i}\ <\ t_{out}\right \}$.

\subsection{The SNN process temporal pulses signal}

\begin{figure}[ht]
\centering
\label{fig:implementation}
\includegraphics[width=1\linewidth]{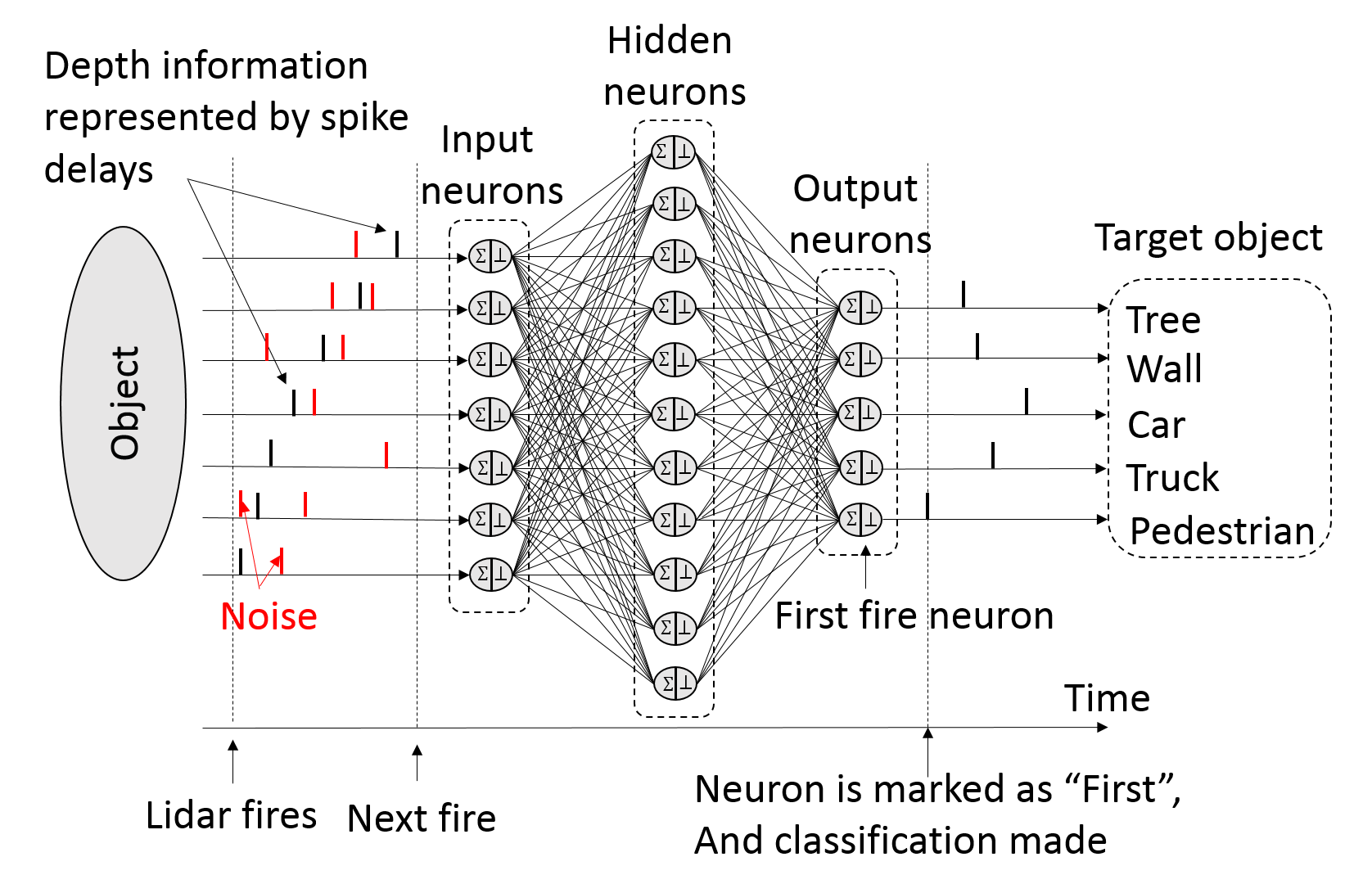}
\caption{The implementation of object detection based on Spiking Neural Network.}
\end{figure}

SNN based on temporal coding could implement temporal pulses signal processing with high accuracy and real time. Firstly, the active function Eq. 3 expresses the relationship between the input spike times and the time of first spike of the output neuron.  Based on this relationship, we could achieve the weights of trained. Input the trained weights to the network as illustrated in Figure \ref{fig:implementation}, for real pulse input, and we set the original time is zero. When the first spike reach to the SNN, it will multiple weight and result is compared with threshold. If the result less than the threshold, the neuron will not fire and cumulate next spike 's result until the sum of result large than threshold. Once the neuron of subsequent layer fire, that neuron will not receive any spike until the network is reset, and a new input pattern is presented. Therefore, when the SNN finish a pattern recognition, it maybe just needs to input a few spikes not all spikes. This allows the SNN to have faster respond. Furthermore, We set the classification through the first fire neuron of output layer, which is beneficial to the acceleration of results. So, the trained SNN could process directly temporal pulses signal from Lidar and implement object detection with high accuracy and real time.

\subsection{Database}

\begin{figure}[ht]
\centering
\label{fig:database}
\includegraphics[width=1\linewidth]{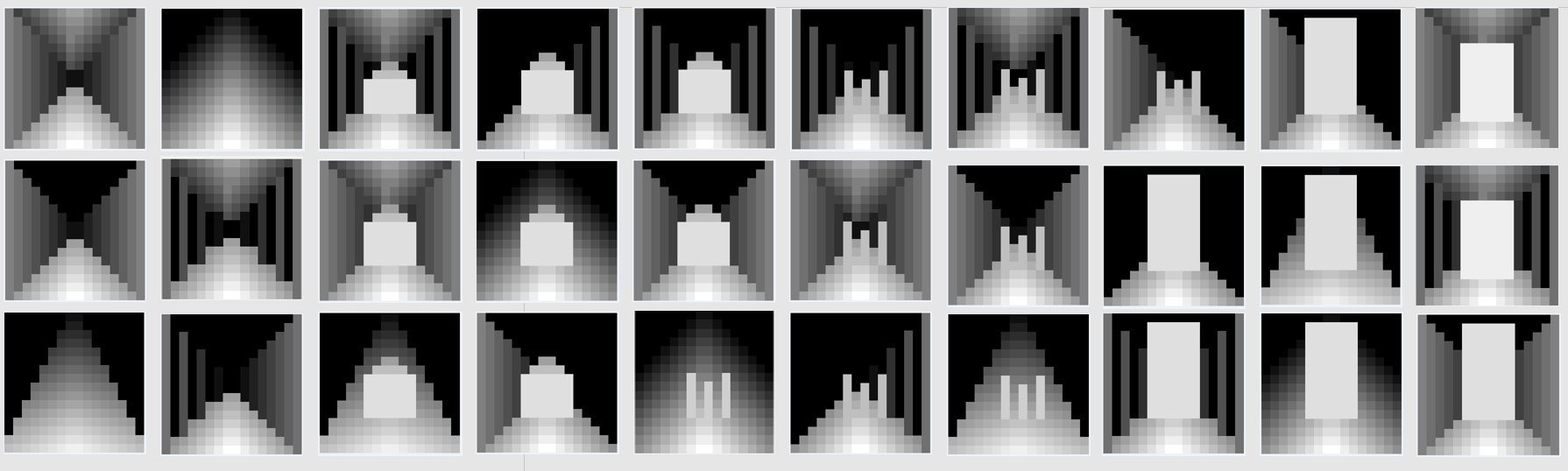}
\caption{The source data-set which include different road condition, car, pedestrians and truck.}
\end{figure}

% \begin{figure}[t]
% 		\begin{subfigure}[normalsize]{0.1\textwidth}
% 			\includegraphics[scale=0.4]{fig/0_1.png}
%         	\caption{$0-0.1$}
%         	\label{fig}
% 		\end{subfigure}
%         \hspace{0.05in}
%         ~
%         \begin{subfigure}[normalsize]{0.1\textwidth}
% 			\includegraphics[scale=0.4]{fig/0_2.png}
%         	\caption{$0-0.2$}
%         	\label{fig}
% 		\end{subfigure}
%         \hspace{0.05in}
%         ~
%          \begin{subfigure}[normalsize]{0.1\textwidth}
% 			\includegraphics[scale=0.4]{fig/0_33.png}
%         	\caption{$0-0.33$}
%         	\label{fig}
% 		\end{subfigure}
%       	\hspace{0.05in}
% 		~
%  		\begin{subfigure}[normalsize]{0.1\textwidth}
% 			\includegraphics[scale=0.4]{fig/0_5.png}
%         	\caption{$0-0.5$}
%         	\label{fig}
% 		\end{subfigure}
%         \caption{A pattern with different range of noise.}
%         \label{fig}
% \end{figure}

\begin{figure}[t]
    \centering
    \subfigure[$0-0.1$]{
        \includegraphics[scale = 0.4]{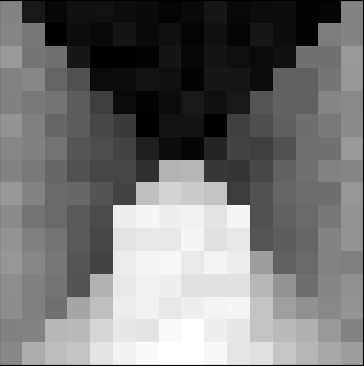}
        \label{fig:noise0.1}
    }
    \hspace{0.05in}
    \subfigure[$0-0.2$]{
        \includegraphics[scale=0.4]{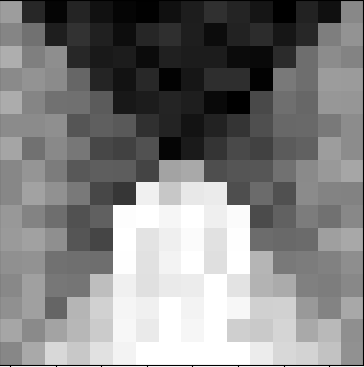}
        \label{fig:noise0.2}
    }
    \hspace{0.05in}
    \subfigure[$0-0.33$]{
        \includegraphics[scale=0.4]{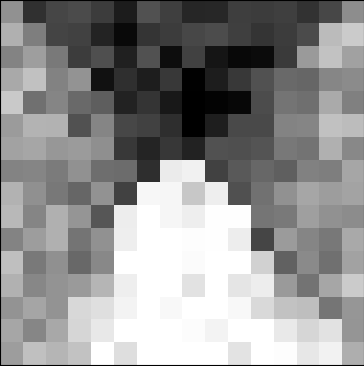}
        \label{fig:noise0.3}
    }
    \hspace{0.05in}
    \subfigure[$0-0.5$]{
        \includegraphics[scale=0.4]{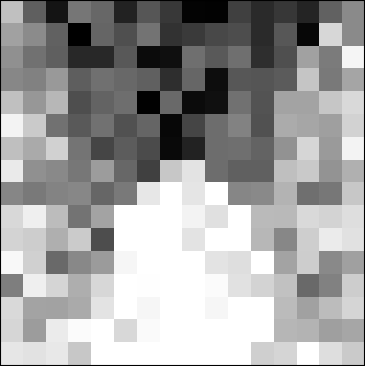}
        \label{fig:noise0.5}
    }
    \caption{A pattern with different range of noise.}
    \label{fig:noise}
\end{figure}

% \begin{figure*}[t]
% 		\begin{center}
% 		\begin{subfigure}[normalsize]{0.47\textwidth}
% 			\includegraphics[scale=0.3]{fig/number2.png}
%         	\caption{Pulses distribution of different noise ranges.}
%         	\label{fig}
% 		\end{subfigure}
%         \hspace{0.1in}
%         ~
%         \begin{subfigure}[normalsize]{0.6\textwidth}
% 			\includegraphics[scale=0.3]{fig/time4.png}
%         	\caption{Recognition time of every pattern.}
%         	\label{fig}
% 		\end{subfigure}
%         \caption{The recognition performance of SNN Object Detection System is reflected by (a) the number of spikes, (b) recognition time}
%         \label{fig}
%         \end{center}
% \end{figure*}

\begin{figure*}[t]
    \centering
    \subfigure[Pulses distribution of different noise ranges.]{
        \includegraphics[scale=0.25]{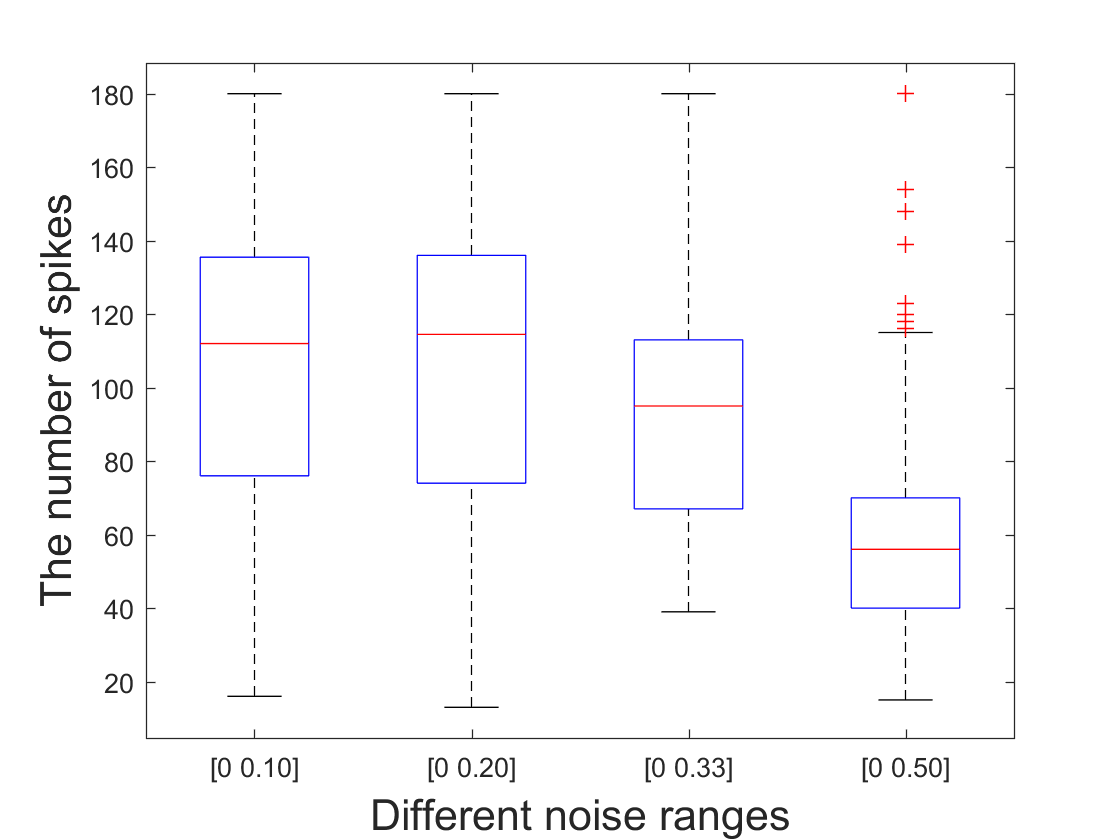}
        \label{fig:performance_spikes}
    }
    % \hspace{0.01in}
    \subfigure[Recognition time of every pattern.]{
        \includegraphics[scale=0.25]{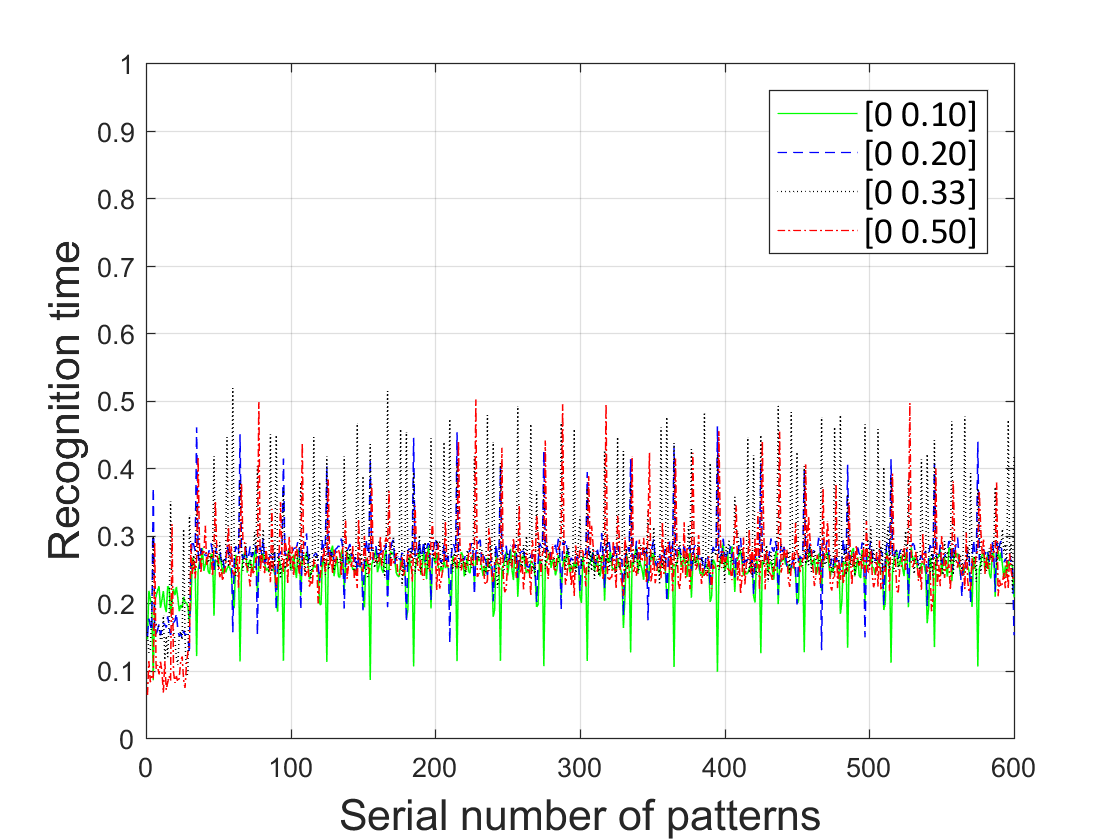}
        \label{fig:performance_time}
    }
    \caption{The recognition performance of SNN Object Detection System is reflected by (a) the number of spikes, (b) recognition time}
    \label{fig:performance}
\end{figure*}

We have to create the complete temporal pulses database due there no such database. The existing Lidar data-set has been pre-processed such as Udacity and KITTI, not the raw temporal pulses from Lidar, so they do not meet our requirements. In order to meet the experimental requirements and actual situation, we use the Velodyne VLP-16 as a simulation Lidar. The Lidar at X time emits some pulses which is reflected by the obstacle, and after these pulses reaches the sensor cell at different delay times. So, these pulses own a time delay property, and form the times delay array. There are two cases: For the same obstacle, different parts of obstacle have different delay times. For different obstacles, farther away from Lidar, delay time more longer and vice-versa. Based on the above rules, the source data is generated through MATLAB and include 30 classes as shown in Figure \ref{fig:database}. The source data from left to right are different road conditions, car, pedestrians and truck. The road conditions include 6 classes: tunnel, road, road with walls, lower bridge, upper bridge, road with wall and street lamp. The target objects include car, pedestrians and truck can combine different road conditions to produce 24 different combinations. The 30 classes correspond to 30 different images. Each image use Lidar's 3D projection to represent test scenario, where Lidar is located at the center of the bottom edge of image and different gray scales represent different time delay. Because the size of image is 16 $\times$ 16, it can not have too many kinds of obstacles. If We put too many kinds of obstacles, it will happen very serious overlap. So we have only designed eight different combinations for each target object. In total, we get 30 classes. On the basis of source data, we can adjust the position information of each class, thus training data and test data can be generated.  Let us think it, the source data consists different delay time, and delay time relate to the type of object and position information. If we add some random values to delay time of the whole object, it will produce a new pattern with delay time array and means different position information of object as well as simulate the dynamic characteristics of object. And then if we add different noises to each pattern, it will produce scenes with different noise. Because of the particularity of the data-set, the data-set generated by this method has three advantages. First of all, each class has a variety of patterns. Next, different noise scenarios are generated. Finally, the test data different from the training data. Therefore, the data-set satisfies the application requirements for testing the performance of our system.

\subsection{Noise}

For the Noise: First of all, the noise here is not a real noise, it simply refers to the object is detected without a label or suspended matter in the air.
\begin{itemize}

\item If the delay time of the noise is greater than that of the object, we do not need to worry about the effect of noise on the classification of object. Because, the network we designed is only processes the first pulse of input, and the latter pulses would not be input into the network. And that's why our method enable ultra-fast information processing, and has higher accuracy. 
\item On condition that the delay time of noise less than the delay time of the object, we have to consider the effect of noise on the classification of objects. For the actual situation, it means that there are some noise that only affect the pulses from the objects. However, the object is not affected. Figure \ref{fig:noise} shows the impact of different range noise on a pattern. From Figure \ref{fig:noise0.1} the range of noise is from 0 to 0.1, and the noise is not a great influence on a pattern. From Figure \ref{fig:noise0.2} the range of noise is from 0 to 0.2, and the influence of noise on a pattern is a little bit. From Figure \ref{fig:noise0.3} the range of noise is from 0 to 0.33, and the influence of noise on a pattern is a lot. From Figure \ref{fig:noise0.5} the range of noise is from 0 to 0.5, and the influence of noise on a pattern is very much. Furthermore, Figure \ref{fig:noise0.3} and \ref{fig:noise0.5} appear the distortion. In the case of noise, the greater of range of noise, the greater the impact on the detection of object. Since the total channel is fixed, as the effect of noise increase, the information from object will decrease. However, how to add the noise to the data-set, we first generated a matrix with uniform distribution values that is same size as a pattern of data-set, and then add the matrix to this pattern. In this process, The matrix is unknown without a label and can not be classified. When we put the matrix to the pattern, the values of pattern will be disturbed and have some values without labels, thereby a new pattern with noise will be generated. Since the uniform distribution of values can be set in different ranges and are random. First of all, For the same range, the every new pattern will be different from each other. Secondly, for different ranges, the new patterns not only own a different range of noise, but also reflect the effects extent of noise on the pattern. Through this method, we can obtain data-sets with different range of noise and test our system’s ability to resist noise.
\end{itemize}

\section{Experimental Results and Analysis}

\subsection{Database and Neuron Network parameter}

SNN with temporal coding is evaluated with object detection from the Database. SNN with temporal coding is relying on Velodyne VLP-16 Lidar. The maximum time delay in SNN algorithm is limited to 1$\mu$s. The database contains 3000 training data-sets and 600 testing data-sets for 30 different categories: Pedestrian, Car, Truck, Building, Bridge and son on. The feedforward network with fully connected has three layers: input layer of 256 neurons, hidden layer of 400 neurons and output layer of 30 neurons. 

\subsection{Training phase}

During training phase, we tried to different learning rate and batch size, when the learning rate is set to 0.01 and batch size is set to 60, the network convergence is the fastest. In addition, the standard stochastic gradient descent (SGD) and max epochs of 100 with L2 regularization was employed for the SNN training. With the increase of epoch, the trend of test accuracy is shown in Figure \ref{fig:accuracy}, and quantitative inference results are provided in Table \uppercase\expandafter{\romannumeral1}. As can be noted from  the table, the average accuracy is up to 99.83\%\ with 10\%\ noise and reduced by about 32.5 percentage point from narrow range (0-0.1) to wide range (0-0.5). When the range of noise is 0 to 0.5 and the scene was badly distorted, the average accuracy of network has been reduced to 68.16\%\. Therefore the maximum anti-noise range of network is from 0 to 0.5.

\begin{table}[h]
\caption{The average accuracy of Object detection in different noise ranges}
\begin{center}
\scalebox{1.2}{
\begin{tabular}{|c|c|}
\hline
Noise range & Average accuracy\\
\hline
[0 0.10] & 99.83\%\\
\hline
[0 0.20] & 96.16\%\\
\hline
[0 0.33] & 82.66\%\\
\hline
[0 0.50] & 68.16\%\\
\hline
\end{tabular}}
\end{center}
\end{table}

\subsection{Object detection under different noise conditions}

\begin{figure}[ht]
\centering
\label{fig:accuracy}
\includegraphics[width=0.6\linewidth]{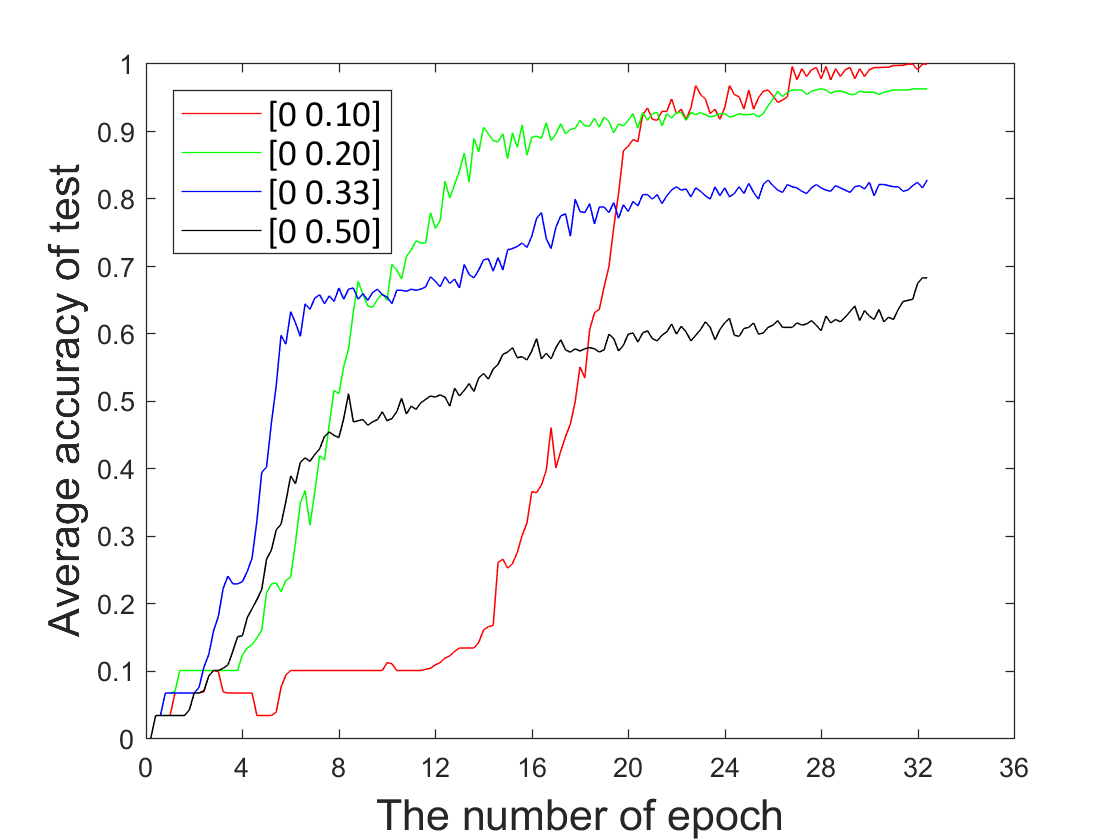}
\caption{The average accuracy of test.}
\end{figure}

We tested four different ranges of noise. From Figure \ref{fig:accuracy} shows that four different average accuracy curves red, green, blue and black correspond to different ranges of noise. At the beginning these four curves have different convergence rates, the latter three curves convergence faster than the first. The fastest convergence range of red curve is from 14 epochs to 20 epochs whereas the other curves are from 1 epoch to 8 epochs. It is proved that noise contributes to the convergence of our network under certain conditions. But it does not mean that the greater the noise range, the faster the convergence. For example the black curve with a noise range of 0 to 0.5 is slower than the blue curve with a noise range of 0 to 0.33. Figure \ref{fig:accuracy} also shows that Even though we have set the maximum number of epochs to be 100, after 32 epochs, the four curves are almost stable and the average accuracy leveled off. Based on the above results, it proves that our network not only has good anti-noise ability but also has a fast convergence speed.

\subsection{Performance Analysis of SNN Object Detection System}

The SNN Object Detection System was evaluated with the recognition time, recognition time distribution and number of spikes based on recognition of 600 patterns. Figure \ref{fig:performance} shows the recognition performance of the SNN Object Detection System. The four boxes in Figure \ref{fig:performance_spikes} represent that the distribution of the number of spikes under different noise condition when the 600 patterns are recognition. The range of noise include 0 to 0.1, 0 to 0.2, 0 to 0.33 and 0 to 0.5 and the number of spikes from 16 to 180, 13 to 180, 39 to 180 and 15 to 180. However, traditional network need to input 256 spikes to complete a pattern recognition comparison with our network maximum input spike number of 180. Figure \ref{fig:performance_time} show that the recognition time and the main distribution of recognition time. The maximum delay time is 1 $\mu$s and the total number of patterns is 600. The four curves in Figure \ref{fig:performance_time} represent the recognition time for each pattern. The four groups correspond to four different noise ranges and three bars display three different time ranges. The height of each bar is the total number of recognition time in the corresponding time range. As seen in Figure \ref{fig:performance_time}, most recognition time are distributed between 0.23 $\mu$s and 0.3 $\mu$s, and the total number of recognition time is 495, 517, 476 and 440 respectively. According the performance of spike number and recognition time, it proved that the SNN Object Detection System can achieved very low latency. 

\section{CONCLUSIONS}

In this paper we introduced the SNN Object Detection System: implement an object detection using Spiking Neural Network with temporal coding and Temporal Pulses from a Lidar perfect combination. We also proved the benefits of the system through quantitative and qualitative experiments. As future works we plan to exploit the networks which deeper than three layers. Detection of more object class can be considered and explored for future works. Moreover, the SNN can be combined with crossbar to implement further acceleration.

\section*{ACKNOWLEDGMENT}

The authors would like to thank the support of Binghamton University.

\bibliographystyle{unsrt}
\bibliography{references}

\end{document}